\documentclass[journal,twoside]{IEEEtran}
\pdfoutput=1
\def \printCitationsAll {}

\def\SECresults{}

\ifx \printNone \undefined
\def\SECintro{}
\def\SECrelated{}
\def\SECcpn{}
\def\SECgraph{}
\def\SECresults{}
\fi

\newcommand{\move}[1]{}
\ifx \printCitationsAll \undefined
\newcommand{\cit}[1]{}
\else
\newcommand{\cit}[2]{\cite{#1}{\color{red}\textbf{#2}}}
\fi

\iftrue

\else

\fi

\usepackage{placeins}
\usepackage[backend=bibtex, backref=true, firstinits=true,sorting=none]{biblatex}
\usepackage[bookmarks=false]{hyperref}

\usepackage{amsmath}
\usepackage{graphicx}
\usepackage[dvipsnames]{xcolor}
\usepackage{epstopdf}
\usepackage{color}
\usepackage{url}
\usepackage{subfig}
\usepackage{multirow}
\usepackage[utf8]{inputenc}

\usepackage{tikz,tkz-graph}
\usetikzlibrary{calc}
\usetikzlibrary{decorations.markings}
\usetikzlibrary{arrows}

\usepackage{algorithmic}
\usepackage[ruled]{algorithm2e}
\ifCLASSINFOpdf
\else
\fi
\begin{document}
%
\title{Cell Tracking via Proposal Generation and Selection}
%
%
%

\author{\IEEEauthorblockN{Saad~Ullah~Akram\IEEEauthorrefmark{1},
Juho~Kannala,
Lauri~Eklund,
and~Janne~Heikkilä}
\thanks{S.U.~Akram is with the Center for Machine Vision and Signal Analysis, and Biocenter Oulu, University of Oulu, Oulu, Finland (e-mail: \href{mailto:saad.akram@oulu.fi}{saad.akram@oulu.fi}).}
\thanks{J.~Kannala is with the Department of Computer Science, Aalto University, Espoo, Finland.}
\thanks{L.~Eklund is with the Faculty of Biochemistry and Molecular Medicine, Biocenter Oulu, and Oulu Center for Cell-Matrix Research, University of Oulu, Oulu, Finland.}
\thanks{J.~Heikkilä is with Center for Machine Vision and Signal Analysis, University of Oulu, Oulu, Finland.}

}

%
%

\markboth{v1.01}%
{Cell Tracking via Proposal Generation and Selection}
%



\maketitle

\begin{abstract}
Microscopy imaging plays a vital role in understanding many biological processes in development and disease.
The recent advances in automation of microscopes and development of methods and markers for live cell imaging has led to rapid growth in the amount of image data being captured.
To efficiently and reliably extract useful insights from these captured sequences, automated cell tracking is essential.
This is a challenging problem due to large variation in the appearance and shapes of cells depending on many factors including imaging methodology, biological characteristics of cells, cell matrix composition, labeling methodology, etc.
Often cell tracking methods require a sequence-specific segmentation method and manual tuning of many tracking parameters, which limits their applicability to sequences other than those they are designed for.
In this paper, we propose 1) a deep learning based cell proposal method, which proposes candidates for cells along with their scores, and 2) a cell tracking method, which links proposals in adjacent frames in a graphical model using edges representing different cellular events and poses joint cell detection and tracking as the selection of a subset of cell and edge proposals.
Our method is completely automated and given enough training data can be applied to a wide variety of microscopy sequences.
We evaluate our method on multiple fluorescence and phase contrast microscopy sequences containing cells of various shapes and appearances from ISBI cell tracking challenge, and show that our method outperforms existing cell tracking methods.
Code is available at: \href{https://github.com/SaadUllahAkram/CellTracker}{https://github.com/SaadUllahAkram/CellTracker}.
\end{abstract}


%
\IEEEpeerreviewmaketitle

\ifdefined\SECintro
\section{Introduction}
\label{sec:intro}
Cell proliferation, division, differentiation, and migration are central to many fundamental biological processes, which include immune response, wound healing, tissue homeostasis, morphogenesis and spread of many diseases including cancer \cite{Trepat2012}. 
In cancer research, cell tracking can shed light on how the cancer cells spread and help in the development of treatments.
In drug discovery research, cell tracking can be used to quantitatively analyze the changes induced by the drug to cell motion and morphology, which can indicate the clinical potential of the given drug \cite{Zimmer2002}.
In developmental biology, cell tracking can create cell lineage trees for complex organisms and organotypic cultures (e.g. zebrafish, fruitfly, kidney etc) which can provide useful insights into how embryos and organs develop, and how embryonic cells differentiate into specialized cells \cite{Amat2014}.


Often the biological processes being investigated are complex and are influenced by hundreds of genes and proteins \cite{Zimmer2006}, which makes it difficult to understand these processes from the study of few cells or samples. 
To extract meaningful insights, it is necessary to analyze large number of cells in multiple samples.
In recent decades, automation of microscopes has enabled the acquisition of time-lapse images capturing multiple spatial locations in a large sample \cite{Hilsenbeck2016} and multiple samples \cite{Neumann2010} from a large-scale experiment \cite{Eliceiri2012}. 
E.g. single experiments in developmental biology capturing developing embryos can generate multiple TBs of image data per hour \cite{Liu2016}.
Manual analysis of this data is extremely time consuming, and produces results which are non-reproducible, often qualitative and limits the hypothesis which can be tested.
All these factors have increased the importance of automated methods for cell tracking.
In recent years, automated analysis has also provided insights which were not feasible with manual analysis either due to scale \cite{Neumann2010} or subtlety \cite{Zimmer2006}.
It has enabled the discovery of periodic membrane deformation of crawling amoeba \cite{Zimmer2006} and the use of high-throughput screening ($\approx$ 190,000 sequences) to identify genes involved in cell division, migration and survival among other biological functions \cite{Neumann2010}.

Cell tracking in general is a very challenging problem due to low signal to noise ratio, poor staining, variable fluorescence in cells or cell organelles, low contrast, high cell density, deformable cell shapes, low inter-cellular shape and appearance variation, and sudden changes in motion direction and speed.
The difficulty of cell tracking depends heavily on the imaging methodology, biological characteristics of living cells, cell matrix composition and cell density. 
Often there are competing constraints, which result in different challenges, e.g., in fluorescence microscopy, imaging at a high frame rate can alleviate some tracking challenges as the change in position and shape of cells reduces but repeated exposure to light can damage cells (photodamage) \cite{Hilsenbeck2016} and alter their behavior or lead to loss of fluorescence (photobleaching) \cite{Meijering2009} making it difficult to detect and track cells as time progresses.

Robust cell detection/segmentation is critical for accurate cell tracking as often more than half of the tracking errors can be traced back to detection errors \cite{Matula2015}.
Often some image regions in a sequence can be very ambiguous and even a trained biologist may have difficulty in correctly detecting all cells.
In most of these regions, few neighboring frames can resolve the ambiguities.
Cell proposals allow consideration of multiple competing hypothesis for these challenging image regions by not making a hard detection decision.
Tracking stage can then consider the temporal information, which often resolves most of these ambiguities and makes temporally consistent decisions leading to more accurate cell tracks.

Many automated cell tracking methods have been proposed in the past \cite{Hilsenbeck2016,Eliceiri2012,Meijering2012b,Maska2014,Held2015}, which achieve very good tracking performance on specific sequences but this problem is still far from solved due to large variation in the challenges involved in the sequences.
Often the tracking methods are designed for a particular set of sequences, tested on private datasets using different evaluation metrics, which makes it difficult to compare these methods with each other.
To address this problem, recently ISBI cell tracking challenges were organized \cite{Maska2014}, which provided a common set of sequences and a standard evaluation protocol, enabling comparison of different tracking methods.



In this paper, we present a joint cell detection and tracking method based on the idea of proposal selection \cite{Engin2015}, which achieves better performance than existing cell tracking methods on multiple fluorescence and phase contrast microscopy datasets from ISBI cell tracking challenge\footnote{\url{http://www.codesolorzano.com/Challenges/CTC/Datasets.html}}.
Our main contributions are: 1) extensions to cell proposal network \cite{Akram2016c} which improve individual cell segmentation masks. 2) a simple framework for cell tracking based on the idea of proposal selection \cite{Engin2015} which can handle different cellular events (mitosis, death, enter, exit and move).
We show that the proposals generated by our convolutional neural network (CNN) are better than those from previous methods and lead to improvement in tracking performance.

Our joint detection and tracking method has three main components: cell proposals (Section~\ref{sec:props}), graphical model (Section~\ref{sec:graph}) and optimization (Section~\ref{sec:solve}).
\fi

\ifdefined\SECrelated
\section{Related Work}
\label{sec:relatedwork}
\subsection{Cell Proposals}
In recent years, object proposals have become very popular in general object detection and segmentation, with almost all leading detection and segmentation methods relying on object proposals \cite{Hosang2015}.
The advantage of using proposals is that they provide a small set of regions likely to contain the objects of interest, which allows the use of computationally expensive features, classifiers and inference for detection, segmentation and tracking.
The challenges in biomedical image analysis are different from general object detection and many object proposal generation methods do not work well when applied directly.
As a result, most cell proposal methods so far have used relatively simple proposal generation stages.
Cell proposal generation methods can be grouped into four main categories: multiple settings \cite{Liu2014,Arteta2012,Bise2015,Jug2014,Arteta2015}, shape \cite{Engin2015,Akram2016a}, super-pixel merging \cite{Schiegg2014, Funke2015} and deep learning \cite{Akram2016b,Akram2016c,Yurchenko2016}.


Methods in first category use multiple settings of a segmentation algorithm to generate cell candidates; \cite{Bise2015,Arteta2015} use multiple thresholdings, \cite{Liu2014} use marker controlled watershed with multiple H-minima values, and \cite{Jug2014} use graph-cut with multiple different unary costs.
Thresholding based methods (e.g. MSER) \cite{Arteta2012, Bise2015} exploit the fact that cell centers are usually brighter than their boundaries and there often exists some optimal threshold at which individual cells can be segmented as separate proposals.
This assumption does not hold true in more challenging regions, where there might not exist a suitable threshold which can separate all touching cells in a cluster.
Allowing each proposal to contain more than one cell \cite{Arteta2015} and/or transforming the images \cite{Arteta2015} so that the above assumption becomes true are two ways of increasing the generalization of these methods.

Shape based methods exploit the fact that cells, especially nuclei, have round or elliptical shapes by using either multi-scale blob detection \cite{Akram2016a} or multi-scale ellipse fitting \cite{Engin2015} to detect and segment cells.

Super-pixel merging methods \cite{Schiegg2014, Funke2015} assume that the intensity change or cell probability change at cell borders is stronger than within cell bodies.
These methods first generate over-segmented super-pixels such that each super-pixel contains pixels from just one cell.
Then, they merge these super-pixels hierarchically to obtain cell proposals.
These methods can handle arbitrary shapes, but they need a criteria for merging superpixels, which can be challenging due to the presence of strong gradients inside cell bodies and weak gradients between touching cells.

Deep learning based methods use a convolutional neural network (CNN) to either regress a cell bounding box \cite{Akram2016b} or represent each image patch with a descriptor, which is used to retrieve the proposal from a cell gallery \cite{Yurchenko2016}.
\cite{Akram2016c} extend \cite{Akram2016b} by adding a second CNN, which outputs masks for each proposed box.

Most cell proposal generation methods do not provide a natural way of ranking or scoring the proposals, so a second stage is used to extract appearance and shape features from each proposed region and these features are used to score it.
The features used are usually hand crafted and consist of basic appearance and shape statistics, including area, mean intensity \cite{Schiegg2014}, histogram of proposal boundary \cite{Arteta2012}, etc.
The classifiers used to score proposals include random forest \cite{Schiegg2014}, gradient boosted trees \cite{Engin2015} or support vector machines \cite{Akram2016a, Bise2015, Arteta2012}.

\subsection{Cell Tracking}
A large variety of cell tracking methods have been proposed over last couple of decades \cite{Hilsenbeck2016,Eliceiri2012,Meijering2012b,Maska2014,Held2015}.
These methods can be grouped into two broad categories: tracking by assignment and tracking by model evolution.
In tracking by model evolution, a mathematical model of cells represented either parametrically \cite{Dufour2011,Amat2014} or implicitly \cite{Dzyubachyk2010} is evolved from frame to frame to jointly segment and track cells.
Parametric models represent cell boundaries using active meshes \cite{Dufour2011}, active contours \cite{Zimmer2002}, or Gaussian Mixture Models (GMM) \cite{Amat2014}.
These methods are computationally fast, their performance depends heavily on the chosen parameterization, and they require separate steps for handling topological changes, i.e cell divisions and cells touching. 
Implicit methods represent the cell contours as the levelset of a function and can naturally handle topological change but are computationally expensive.
Model evolution methods typically require high frame rate, high resolution and usually need special handling of cells entering or leaving the imaged region.

Tracking by assignment is a much more popular approach within cell tracking field with five out of six methods in ISBI 2013 Cell Tracking Challenge \cite{Maska2014} belonging to this category.
It decouples cell detection and cell tracking, which makes it more generalizeable and easier to apply to different sequences.
In the first detection stage, cells are detected and segmented using a sequence-specific method and then some features are extracted to represent shape and appearance of each detected cell.
In the second tracking stage, detected cells in neighboring frames are linked with each other.
Cell tracks can then be obtained by considering the whole graph with all frames linked together \cite{Magnusson2012a, Lou2014a} or just neighboring frames \cite{Padfield2011} or a combination of both of these \cite{Kanade2011}.
These cell tracks are more robust when large temporal context is used which ensures that the found tracks have better consistency and are more tolerant to detection errors in individual frames.
Earlier methods were not able to correct cell detection errors in the tracking stage.
However, recent methods \cite{Bise2011,Schiegg2013,Magnusson2015,Haubold2016} are able to resolve false positives, false negatives and under-segmentation errors in the tracking stage. 
To resolve these errors, these methods have to introduce additional complexities in their methods and they can still fail to correct some detection errors.

Recently, few cell proposals based joint cell detection and tracking methods have been presented \cite{Jug2014, Schiegg2014, Engin2015} to utilize temporal information when performing joint detection and tracking.
These methods generate a large set of cell proposals covering multiple hypothesis for ambiguous regions; connect proposals in a spatio-temporal graph to prevent selection of conflicting proposals; and then, use integer linear programming for inference.
Sequence-specific constraints can be easily incorporated in these tracking graphs to improve tracking performance, e.g. \cite{Jug2014} use special exit constraints specific to their particular application and \cite{Schiegg2014} use high level cues about the number of cells in each proposal tree to improve their performance.
These methods also enable easy incorporation of user feedback to correct tracking errors, e.g. a user can mark the number of cells in an under-segmented region and a new constraint fixing the number of cells in that region can be introduced, which can potentially fix the detection/tracking errors locally.
These proposal based tracking methods can be computationally more expensive than the detection based methods as the number of proposals and edges linking them can be considerably higher.
If the proposal quality is very good, then this additional computational requirement might not be very significant.
Otherwise, possible transitions between frames can be restricted to speed up inference for long sequences \cite{Schiegg2014} which might not always be suitable especially if sequences have low frame rate and high cell density. 

\fi
\ifdefined\SECcpn
\section{Cell Proposals}
\label{sec:props}
\label{sec:cpn}

\begin{figure*}[!htb]
\centering
\includegraphics[width=0.95\textwidth]{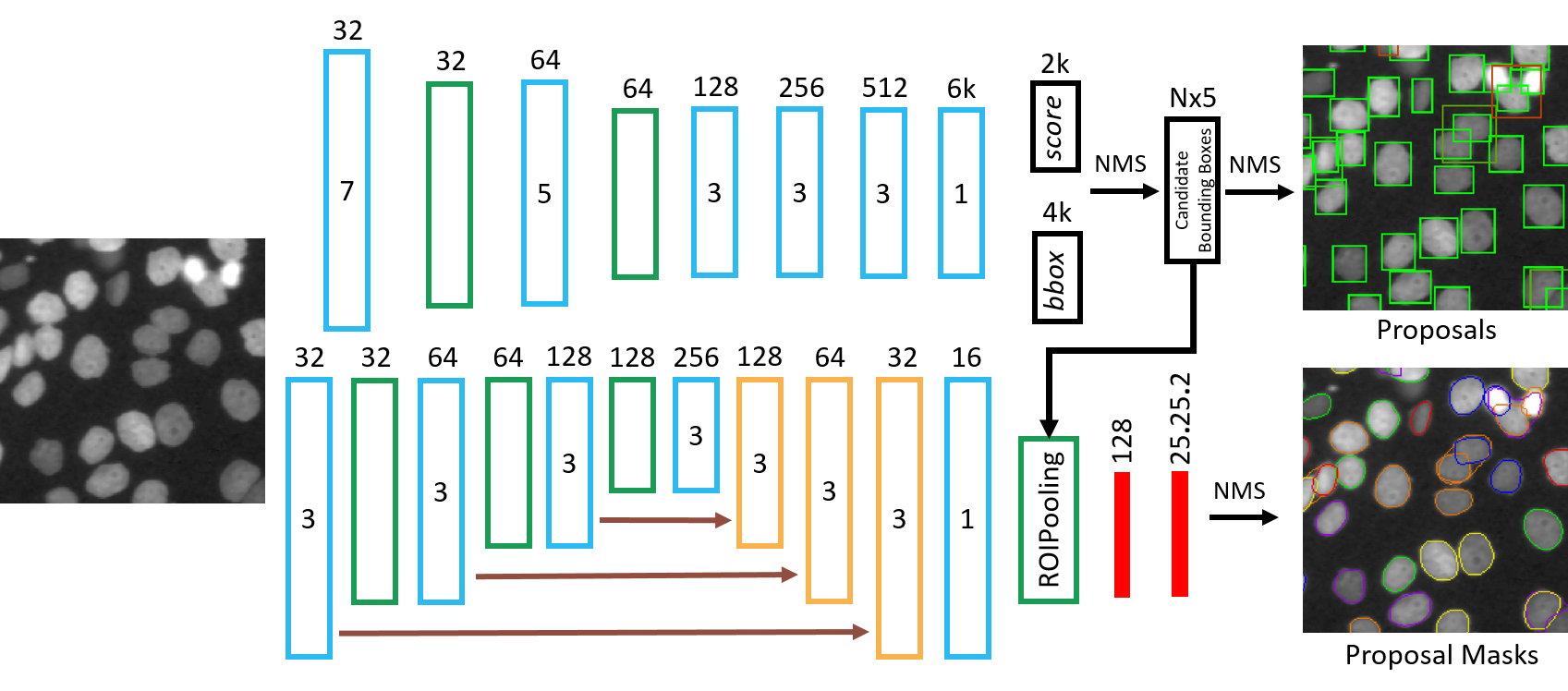}
\caption{Cell Proposal Network (CPN): Top half shows the first network, which proposes \textit{N} bounding boxes and their scores.
Bottom half shows the second network which generate segmentation masks for the \textit{N} proposals.
{\color{ProcessBlue}Convolutional (blue)} (filter size inside the box), {\color{ForestGreen}max-pooling (green)}, {\color{red}fully connected (red)}, and {\color{YellowOrange}deconvolutional (orange)} layers, with the number of feature maps on top of each layer, are shown.
{\color{Brown}$\rightarrow$ } indicates which feature maps are combined by summation.
Proposed bounding boxes and segmentation masks after non-maxima suppression (NMS) are shown for a selected area from \textit{Fluo-N2DL-HeLa} dataset.
Box color indicates proposal score, with bright green representing high score and bright red representing low score.}
\label{fig:nets}
\end{figure*}

The main goal of cell proposal generation is to produce a set of cell candidates, which has a very high recall for individual cells with as few cell candidates as possible.
Each proposal should also have a probability associated with it indicating how likely it is to be an actual cell.

Recently, region proposal network (RPN) \cite{Rena} was proposed to provide object candidates for general object detection and has been extended to generate object segmentation masks \cite{Dai2015b,He2017} and improved upon to better utilize multiple feature maps for proposing better candidates \cite{Lin2016}.
RPN has been previously applied to microscopy images to propose cell candidates \cite{Akram2016b,Akram2016c}.

Here, we present an extension of convolutional neural network (CNN) based cell proposal generation method \cite{Akram2016c}.
Our main extensions are: 1) Use of fully connected layers after ROI-pooling to improve the network's ability to ignore parts of other cells. 2) A gradual merging of feature maps from different network depths to reduce the computational cost and memory requirement.

Our proposal generation method consists of two networks.
First network, shown in the upper half of Fig.~\ref{fig:nets}, proposes cell candidate bounding boxes and their scores.
Second network, shown in the bottom half of Fig.~\ref{fig:nets}, proposes cell segmentation masks for each proposed candidate bounding box.

\subsection{Bounding Box Network}
We use a fully convolutional neural network to propose candidates for cell bounding boxes and their associated scores - probability of them being cells.
This network is based on ZF model \cite{Zeiler2014}; we have reduced the stride in all convolutional layers to 1 as the cells are small compared to general objects and have modified the number of feature maps to reduce the chances of over-fitting.
First five convolutional layers of this network extract a 512-dimensional feature vector (\textit{conv5}) from 45x45 rectangular windows in a sliding window fashion with combined stride of 4 in the input image.
Then, two fully connected layers, \textit{score} and \textit{bbox}, implemented as 1x1 convolutional layers predict cell candidate bounding boxes and scores for these boxes respectively.
\textit{score} and \textit{bbox} layers slide over \textit{conv5} feature map and predict \textit{k} bounding boxes and \textit{k} scores for each pixel in \textit{conv5}.
To predict multiple boxes of varying sizes and aspect ratios from same feature vector, we use anchors \cite{Rena}.
Anchors are reference bounding boxes centered in the input image at the center of receptive field of each \textit{conv5} feature map pixel.
We use 9 anchors for each network; anchor sizes are selected by clustering ground truth (GT) cell bounding boxes using k-means clustering.
The outputs of bbox layer, $b_i$, are the parameters of predicted bounding boxes, $b=(x,y,w,h)$, relative to anchor bounding boxes, $b_a=(x_a,y_a,w_a,h_a)$ \cite{Rena}.
\begin{equation}
b_{i} = ((x-x_a)/w_a, (y-y_a)/h_a, log(w/w_a), log(h/h_a))
\label{eq:bb_para}
\end{equation}

This network proposes bounding boxes very densely and there may be multiple boxes for one cell; we use non-maxima suppression (with intersection over union overlap (IoU = 0.8)) to get rid of most duplicate proposals.

All convolutional layers in this network use a stride of 1 pixel and no padding.
ReLu non-linearities followed by local contrast normalization with same normalization parameters as ZF model \cite{Zeiler2014} are used after each of the first four convolutional layers.
Only ReLu non-linearities are used after fifth convolutional layer.
Both max-pooling layers use stride of 2, padding of 1 and filter size of 3x3.

\textbf{Training:} We train this network using positive and negative samples, which are generated based on intersection over union overlap (IoU) of anchors with ground truth (GT) cell bounding boxes.
Two types of anchors are used as positive samples: first, anchors which have overlap greater than a threshold (0.5); second, anchors having the highest overlap with each GT cell.
Second condition is needed to obtain a positive sample for cells with considerably different size compared to the chosen anchors.
Negative samples are sampled from anchors having overlap with GT cells below a threshold (0.4).
Anchors which are neither labeled as positive or negative are not used during training.
The multi-task loss function used for training this network is \cite{Rena}:
\begin{equation}
L(p_i, b_i) = L_{score}(p_i, p_i^*) + \lambda  p_i^*L_{bbox}(b_i, b_i^*)
\label{eq:loss}
\end{equation}
where the first component $L_{score}$ is soft-max log-loss for two classes, cell and background, and the second component ${L_{bbox}}$ is smooth-$L_1$ loss \cite{Girshick2015}, which penalizes differences between predicted ($b_i$) and ground truth ($b_i^*$) bounding box parameters.
$p_i$ is the probability of the proposed bounding box, $b_i$, being a cell.
Only positive samples ($p_i^* = 1$) contribute to the $L_{bbox}$.
$\lambda$ (10) is a parameter which balances the bounding box regression loss relative to the classification loss.

The weights of this network are initialized randomly from a Gaussian distribution with zero-mean and 0.01 standard deviation.
Network is trained using momentum of 0.9 and learning rate of 0.001 for first 30k iterations, then learning rate is reduced to 0.0001 for next 10k iterations.

We use rotations and flips to augment training data.

\subsection{Segmentation Network}
Our segmentation network takes two inputs: an image and \textit{N} candidate bounding boxes proposed by our bounding box network; and outputs \textit{N} segmentation masks, one for each candidate bounding box.
The network structure has similar shape as U-Net \cite{Ronneberger2015} but has fewer layers as we are working with smaller cells and fewer number of feature maps in the existing layers to reduce the chances of overfitting.

This network consists of two parts; first part contains five convolutional layers, first four of which extract four feature maps (\textit{conv1}, \textit{conv2}, \textit{conv3} and \textit{conv4}) from different depths of the network capturing both low level information necessary for accurate cell boundary localization and high level coarse information needed for ignoring parts of other cells in the proposed bounding box.
The feature maps (\textit{conv2}, \textit{conv3} and \textit{conv4}) are gradually up-scaled using deconvolutional layers and merged by summing them as shown in Fig.~\ref{fig:nets}.
This produces feature maps of same resolution as the input image, the dimensionality of which is reduced by the fifth and final convolutional layer from 32 to 16.
The \textit{conv5} feature maps are shared across all candidate bounding boxes.

The second part extracts fixed sized feature maps for each bounding box and uses it to predict the segmentation mask.
The cell candidate bounding boxes can have varying sizes, so region of interest (ROI) pooling \cite{Girshick2015} is used to adaptively max-pool the region of \textit{conv5} which is inside a given bounding box to a fixed size (25x25).
Then, two fully connected layers are used for predicting 25x25 segmentation mask, which is resized back to the original bounding box size using bicubic interpolation, thresholded (0.5) and the largest connected component is used as the segmentation mask.
After obtaining masks for all proposals, we use non-maxima suppression (with mask overlap (IoU = 0.7)) to remove duplicate proposals. 

All convolutional layers in this network use a stride of 1 pixel and appropriate padding to preserve feature map size.
ReLu non-linearities followed by local contrast normalization with same normalization parameters as ZF model \cite{Zeiler2014} are used after all but last convolutional layer.
All three max-pooling layers use stride of 2, padding of 1 and filter size of 3x3.
Dropout layer with dropout probability of 0.5 is used after the first fully connected layer to reduce the chances of overfitting.

\textbf{Training:} This network is trained using the ground truth (GT) bounding boxes and the bounding box candidates from the bounding box network, which have overlap greater than a threshold (0.6) with any GT cell bounding box.
For each training box, a 25x25 binary segmentation mask is used as the target output during training.
This mask is obtained by cropping the part of GT segmented image which lies inside the candidate bounding box, resizing it to 25x25 using nearest neighbor interpolation and labeling pixels of the largest cell within that box as foreground and rest as background.
The candidate bounding boxes can have small localization errors, which can result in clipping some parts of cells \cite{Akram2016c}.
To resolve this issue, we expand each bounding box by 3 pixels on each side.
The loss function used to train this network is a pixel-wise soft-max log-loss.

All convolutional layers are initialized by randomly sampling from a uniform distribution $U[-\sqrt{3/n}, \sqrt{3/n}]$ \cite{Glorot2010}, where $n$ is the number of incoming nodes.
Deconvolutional layers are initialized using coefficients of bilinear interpolation and fully connected layers are initialized by randomly sampling from a Gaussian distribution with zero-mean and 0.001 standard deviation.
Network is trained using momentum of 0.99 and learning rate of 0.0001 for first 30k iterations, then learning rate is reduced to 0.00001 for next 10k iterations.
Training data is augmented using rotations and flips.

\subsection{Proposal Conflicts:}
Multiple cell proposals can have some pixels in common.
During tracking, it is important that the set of proposals selected does not contain any pair of proposals which conflict, i.e., have substantial overlap with each other.
If the proposals are generated in a hierarchical manner then any overlap between them can be considered as a conflict \cite{Schiegg2014,Engin2015}.
However, when proposals are generated independently, it is unavoidable for some proposals representing different cells to have small overlap.
We use the following criteria to determine, if a proposal $p_i$, conflicts with another proposal, $p_j$:
\begin{itemize}
\item $p_i$ has large overlap with $p_j$, $\frac{|\mbox{p}_i \bigcap \mbox{p}_j|}{|\mbox{p}_i \bigcup \mbox{p}_j|} > C_1$.
\item $p_i$ is completely or mostly inside $p_j$, $\frac{|\mbox{p}_i \bigcap \mbox{p}_j|}{|\mbox{p}_i|} > C_2$.
\item $p_j$ is completely or mostly inside $p_i$, $\frac{|\mbox{p}_i \bigcap \mbox{p}_j|}{|\mbox{p}_j|} > C_2$.
\end{itemize}
where $C_1$ and $C_2$ are the thresholds.

\fi

\ifdefined\SECgraph
\section{Graphical Model}
\label{sec:graph}

\begin{figure}[!t]
\centering
    \subfloat[\label{subfig:tra_graph}]{
      \includegraphics[width=0.45\textwidth]{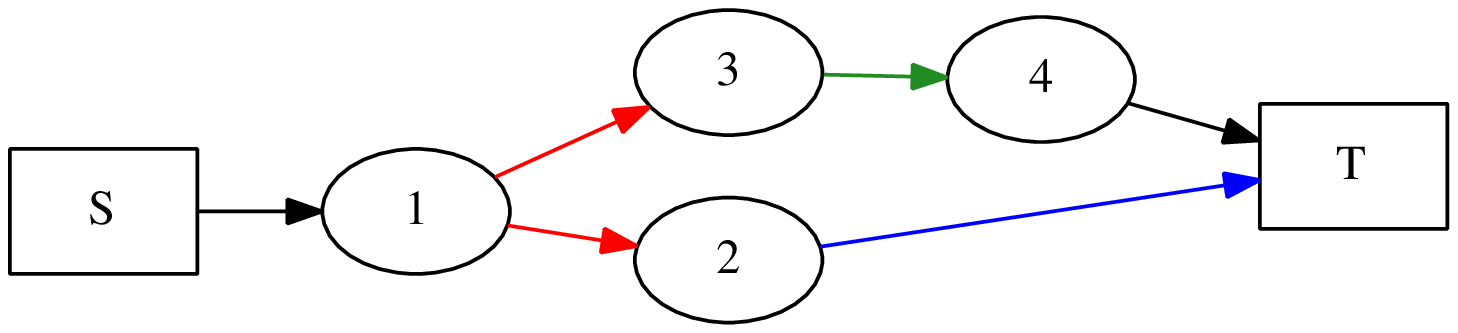}
    }\\
    \subfloat[\label{subfig:tra_prop_graph}]{
      \includegraphics[width=0.45\textwidth]{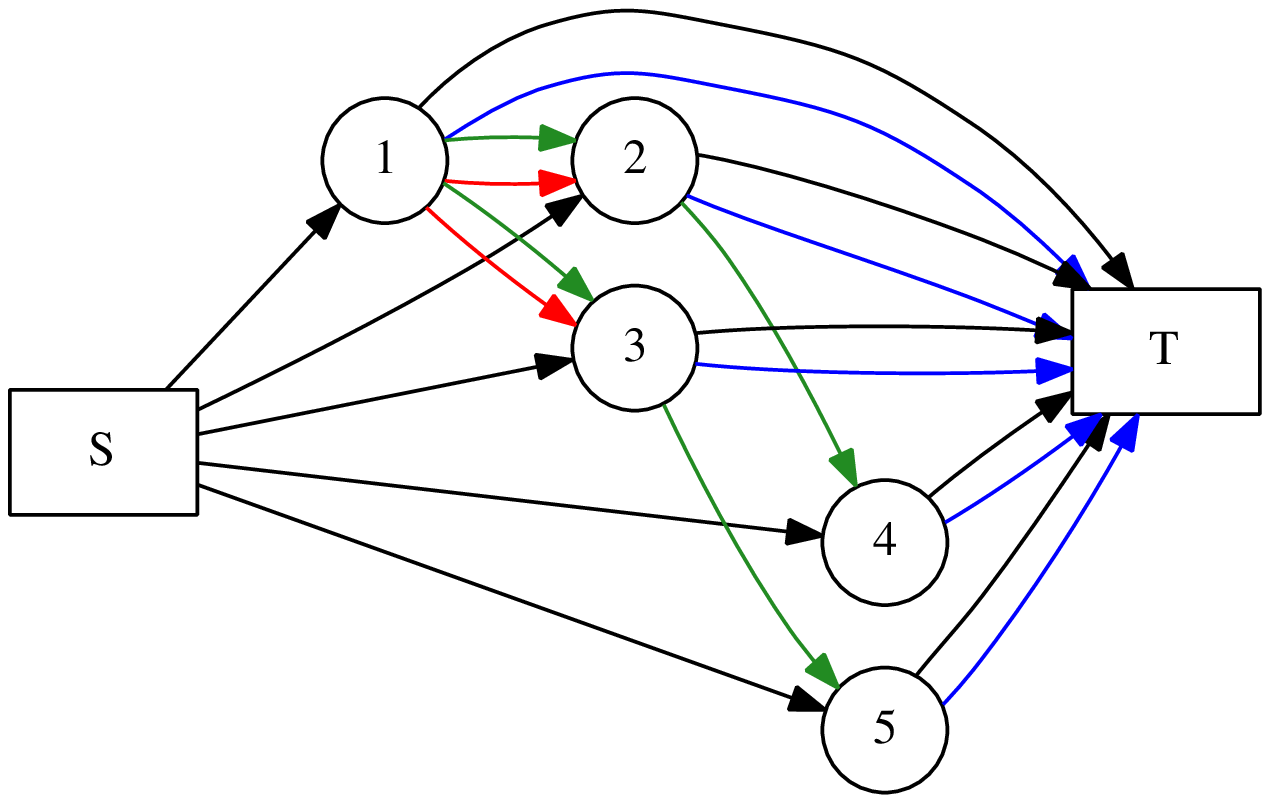}
    }
 \caption{a) Ground Truth (GT) graph ($g^{GT}$) showing 4 cells and one event of each type: {\color{red}mitosis (red)}, {\color{blue}death (blue)}, enter (black), exit (black) and {\color{ForestGreen}move (green)}.
b) A proposal graph ($G$) constructed from 5 proposals, which contains the $g^{GT}$ as one of its sub-graphs.}
\label{fig:graph}
\end{figure}

\subsection{Graph Structure}
Cell tracks can be represented as a directed acyclic graph, $g^{GT} = (N_{g^{GT}}, E_{g^{GT}})$, where $E_{g^{GT}}$ is the set of edges representing different cellular events and $N_{g^{GT}} = \{P_{g^{GT}}, S, T\}$ is the set of nodes in the graph, where $P_{g^{GT}}$ is the set of nodes representing all cells in the sequence.
Two special nodes, $S$ (Source) and $T$ (Sink), are used to represent initiation of new tracks for cells which enter the imaged region and to terminate existing tracks of cells which move out of the imaged region or disappear due to cell death.
Fig.~\ref{subfig:tra_graph} shows a tracking graph for a sequence of 3 frames containing 4 cells and one event of each type: enter, move, exit, death and mitosis.
If the edges between $S$, $T$ and cell nodes are removed, this graph has similar structure as a forest, with each tree in the forest representing the lineage of one cell in the sequence.

To represent all cellular events in a sequence, three types of edges are needed, $e^{\rightarrow}_{i,j}$, $e^{\bullet}_{i,T}$ and $e^{\div,k}_{i,j}$.
Edge $e^{\rightarrow}_{i,j}$ represents the event that a cell moves from node $i$ to node $j$.
There are two special cases of this move edge; edge $e^{\rightarrow}_{S,j}$ represents the entry of a new cell $j$ from outside the imaged region and edge $e^{\rightarrow}_{i,T}$ represents the exit of a cell $i$ from the imaged region.
Edge $e^{\bullet}_{i,T}$ represents the disappearance of a cell due to cell death.
Edge $e^{\div,k}_{i,j}$ represents the link between a parent cell $i$ and a daughter cell $j$ and $k\in\{1,2\}$ is the daughter number.
In normal conditions, a parent cell usually divides into exactly two daughter cells but in some abnormal cases, a parent cell may divide into more than two cells \cite{Tse2012} and in these cases more exit edges from a parent cell can be used to represent the relationship between parent and each daughter cell.
The goal of any cell tracking method is to recover this graph, $g^{GT}$, given a sequence of images.

Cell detections are often noisy and are one major source of errors in tracking \cite{Matula2015}.
To reduce these tracking errors, some detection based methods allow multiple cells to occupy each detection and multiple cells to pass through an edge linking detections \cite{Magnusson2015,Haubold2016}.
Similarly, some methods allow links between non-adjacent frames to recover from false negative errors \cite{Magnusson2015}.
Introduction of these error handling adjustments makes the tracking graph complicated and even with these adjustments, tracking method may not always be able to recover from some detection errors.

Proposal based cell tracking allows a simpler formulation of cell tracking problem as it removes the need for correcting detection errors in the tracking stage.
Given a set of proposals, $p_i \in P$, which have almost perfect cell recall, they can be linked in adjacent frames such that the recall for all cellular events, $e^{c}_{i,j} \in E$, is very high, to create the graph $G$ as shown in Fig.~\ref{subfig:tra_prop_graph}.
Then, the cell tracking graph $g^{GT} = (N_{g^{GT}}, E_{g^{GT}})$ can be recovered almost perfectly from the proposal graph $G = (N_G, E_G)$ by selecting the appropriate proposals and their links.
The cell tracking problem can then be formulated as the selection of the sub-graph $g^*$ which has the highest probability \cite{Engin2015}.
\begin{align}
g^{*} & = \underset{g}{\arg\max} \quad P(g|I)\label{eq:fun11}\\
& \approx \underset{g \subseteq G}{\arg\max} \quad P(g|G)\label{eq:fun12}\\
& =  \underset{g \subseteq G}{\arg\max}(\underset{p_i \in P_g}\prod P(p_i | I_{t(i)})\times\underset{e^{c}_{i,j} \in E_g}\prod P(e^{c}_{i,j} | I_{t(i)}, I_{t(j)}))\label{eq:fun13} 
\end{align}
where $I_{t(i)} \in I$ is the image at time $t(i)$ and $c \in \{\rightarrow, \div, \bullet\}$ is the type (class) of edge.
Eq. (\ref{eq:fun13}) can be obtained from (\ref{eq:fun12}) by assuming that the proposal and event probabilities have markov property, i.e. the probability of a proposal being a cell or an edge linking two nodes does not depend on the previous images given the current images.
In general object tracking this can be a severe limitation but in most cell tracking applications, where cells often abruptly change their movement speed and direction, it does not have major impact on performance.


\subsection{Model Probabilities}
Appearance and shape of each cell proposal is represented using a 92-d feature vector \cite{Arteta2012} consisting of:
\begin{itemize}
\item Histogram (15-d) of proposal intensity.
\item Histogram (8-d) of difference in intensities between the proposal boundary and its dilation for two different dilation radii.
\item Histogram (60-d) of the distribution of the proposal boundary on a size-normalized polar coordinate.
\item Area (1-d) of the proposal.
\end{itemize}

The proposal nodes in the graphical model, \textit{G}, have a cost associated with them, which depends on the likelihood of that proposal being a cell.
Similarly, the weight of an edge in the graph \textit{G} represents the probability of the event represented by that edge.
There are three types of edges representing five different types of events: mitosis, death, enter, leave and move.

The most common and critical type of edge for accurate tracking is the move edge, $e^{\rightarrow}_{i,j}$, which represents a cell moving from a proposal $p_i$ at $t$ to a proposal $p_j$ at $t+1$.
We use gating to restrict edges between proposals which are far away from each other for computational reasons.
Each proposal, $p_i$ at $t$ is linked with proposal $p_j$ at $t+1$ only if $p_j$ is within a fixed radius window from centroid of $p_i$.
This restriction has very little impact on the performance as selected gating threshold is higher than the distance moved by almost all cells.
Each move edge in the training sequence is labeled as positive sample if it connects two proposal, both of which have only one GT marker inside their body and that marker belongs to same cell; otherwise the edge is labeled as negative sample.
The feature set used for classifying move edges consists of:
\begin{itemize}
\item Shape and appearance features extracted from $p_i$ and $p_j$.
\item Distance between $p_i$ and $p_j$.
\item IoU overlap between $p_i$ and $p_j$ both at their original positions and after aligning their centroids.
\item Normalized absolute difference between features of $p_i$ and $p_j$.
\item Probability of $p_i$ and $p_j$ being a cell.
\end{itemize}

A random forest classifier is trained and used to set weight of move edges in the test sequence.

Another important type of edge is mitosis edge, $e^{\div,k}_{i,j}$.
If the image resolution and frame rate are high, then mitosis events can be detected from a single image.
However, in lower resolution and low frame rate sequences, it can be very challenging to detect when a cell is going to divide as the parent cell may be at the start of mitosis event in the last frame before division and as a result it may have similar appearance and shape as non-mitotic cells.
However, there are still some cues, which can help with mitosis detection; these include: firstly, the daughter cells typically are smaller, more elongated and brighter; secondly, both daughter cells often have similar shape and appearance; and thirdly, daughter cells often move in opposite direction so that the line linking them passes close to the parent cell.
By considering the appearance, position and shape of daughter and parent cells jointly, mitosis detection becomes much easier.

We consider $n$ nearest neighbors (with distance from parent below a threshold) of each proposal in the next frame as its potential daughters.
For each combination of parent and daughter pairs (mitosis set), the probability of mitosis is computed and used to set the weight of edges linking the parent to daughter proposals in that set.
Mitosis set, in which each proposal contains only one cell marker and those markers are from a GT parent and its daughters, are used as positive samples, rest are used as negative samples.
The feature set used to train a random forest classifier for scoring mitosis edges consists of:
\begin{itemize}
\item Distance between each pair of member in mitosis set.
\item IoU overlap between each pair of member in mitosis set both at their original positions and after aligning their centroids.
\item Shortest distance of parent centroid from line connecting daughter centroids.
\item Absolute difference of distance between daughter cells and sum of distance between parent and daughter cells.
\item Features of parent and both daughter proposals.
\item Probability of the parent and each daughter being a cell.
\end{itemize}

Since the sequences we are using do not have cell death events marked, those edges are not used.
For sequences with cell death events, a classifier can be trained to predict the likelihood of cell death using the shape and appearance features listed earlier or some other task specific features.

For enter and exit edges, we use a constant probability for each sequence.


\section{Inference}
\label{sec:solve}
\label{sec:ilp}
Cell tracking objective (\ref{eq:fun13}) can be optimized using Integer Linear Programming (ILP) or a slight modification of k-shortest path algorithm (KSP)\cite{Berclaz2011,Haubold2016}.
ILP is able to find a globally optimal solution, which leads to slightly better tracking performance than the approximate but computationally less demanding \cite{Haubold2016} solution found by KSP, especially for sequences with high cell density and frequent cell-cell contact.
In this section, we describe an Integer linear program similar to \cite{Engin2015} that we use to optimize (\ref{eq:fun13}).

Given a proposal graph $G = (N_G, E_G)$, a binary variable $x$ is created for each proposal and edge in the graph $G$.
There are four types of binary variables: first, $x_i^p$ for each proposal $p_i \in P_G$; second, $x_{i,j}^{\rightarrow}$ for each move edge, $e^{\rightarrow}_{i,j} \in E_G$; third, $x_{i,T}^{\bullet}$ for each cell death edge, $e^{\bullet}_{i,T} \in E_G$; and finally, $x_{i,j}^{\div,k}$ for each mitosis edge, $e^{\div,k}_{i,j} \in E_G$, where $k \in \{1,2\}$ is the daughter number.

The cost function minimized by ILP formulation is:
\begin{align}
cost =  \underset{g \subseteq G}{\min}(\underset{p_i \in P_g}\sum x_i^p \cdot cost(p_i) + \underset{e^{c}_{i,j} \in E_g}\sum x_{i,j}^c \cdot cost(e^{c}_{i,j}) )\label{eq:fun_ilp}
\end{align}

To ensure that the ILP finds tracks which are feasible, four types of constraints are required.
First set of constraints are needed to ensure that the selected proposals do not conflict with each other.
Conflicts between selected proposals can be avoided by creating a constraint for each pair of proposals which conflict with each other.
\begin{equation}
x_i^p + x_j^p \leq 1, \quad if \quad C(p_i, p_j) = 1
\label{eq:ilp_constraint2}
\end{equation}
where $C$ is the conflict matrix and $C(p_i, p_j) = 1$ if the proposal $p_i$ and $p_j$ conflict with each other.

Second set of constraints are needed to ensure that one incoming edge is selected for each selected proposal, $p_i$, and no incoming edge is selected if a proposal is not in the found solution.
\begin{equation}
\begin{split}
\sum_{e_{n,i}^c \in E_G} x_{n,i}^c = x_i^p \\
\label{eq:ilp_in}
\end{split}
\end{equation}
where $n \in N_G$ is either a proposal node in adjacent frame or a terminal node ($S$ or $T$).

Third set of constraints are needed to ensure that the number of incoming and outgoing edges is same for each proposal.
\begin{equation}
\begin{split}
\sum_{e_{n,i}^c \in E_G} x_{n,i}^c = \sum_{e_{i,n}^c \in E_G \backslash E_{m_2}} x_{i,n}^c
\label{eq:ilp_out}
\end{split}
\end{equation}
$E_{m_2}$ contains mitosis edges from a parent cell to its second daughter cell, meaning that if a cell goes through mitosis, only its edge to first daughter is considered as the leaving edge.


Fourth set of constraints are needed to ensure that if a mitosis event is selected then edges from parent $i$ to both daughter cells, $m$ and $n$, are selected as well.
\begin{equation}
\begin{split}
x_{i,n}^{\div,1} = x_{i,m}^{\div,2}
\end{split}
\label{eq:ilp_div}
\end{equation}

The tracking integer linear program can be very large with millions of variables, so we use the Gurobi Optimizer\footnote{http://www.gurobi.com/products/gurobi-optimizer} to optimize it.
\fi

\ifdefined\SECresults
\section{Results}
\label{sec:results}

\subsection{Datasets}

\newcommand{\dataw}{0.24}
\begin{figure*}[!t]
\centering
    \subfloat[Fluo-N2DH-HeLa]{\label{subfig:df12}
      \includegraphics[width=\dataw\textwidth]{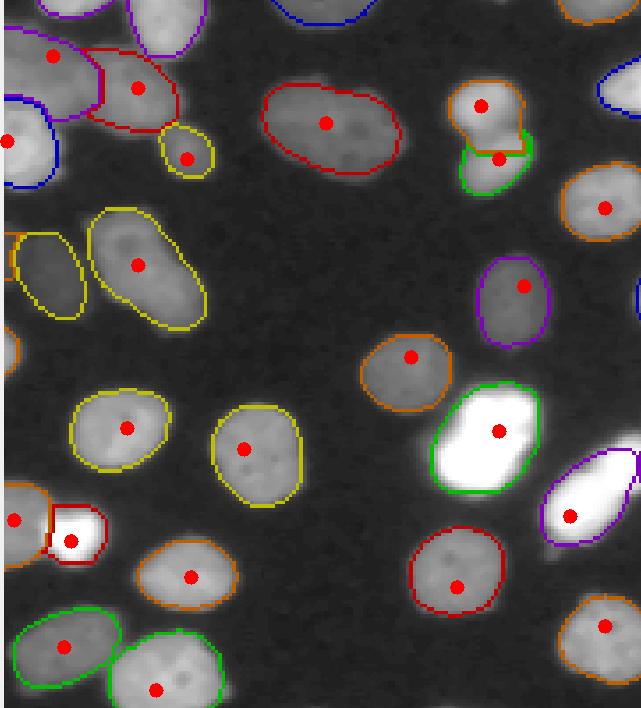}
    }\hspace{-0.75em}
    \subfloat[Fluo-N2DH-GOWT1]{\label{subfig:df22}
      \includegraphics[width=\dataw\textwidth]{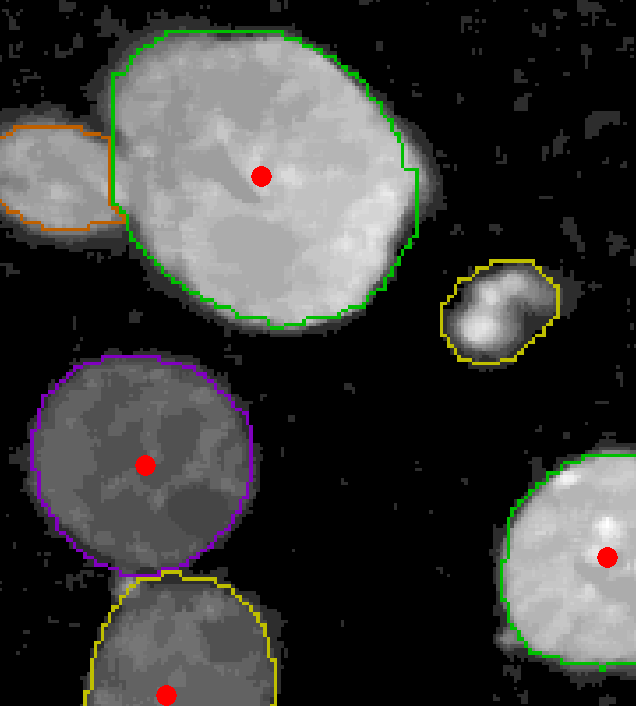}
    }\hspace{-0.75em}
    \subfloat[PhC-C2DH-U373]{\label{subfig:df32}
      \includegraphics[width=\dataw\textwidth]{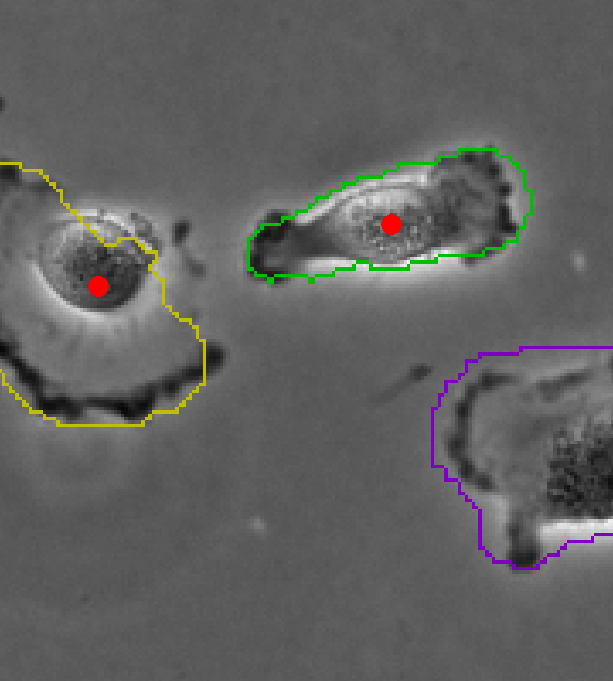}
    }\hspace{-0.75em}
    \subfloat[PhC-C2DL-PSC]{\label{subfig:df42}
      \includegraphics[width=\dataw\textwidth]{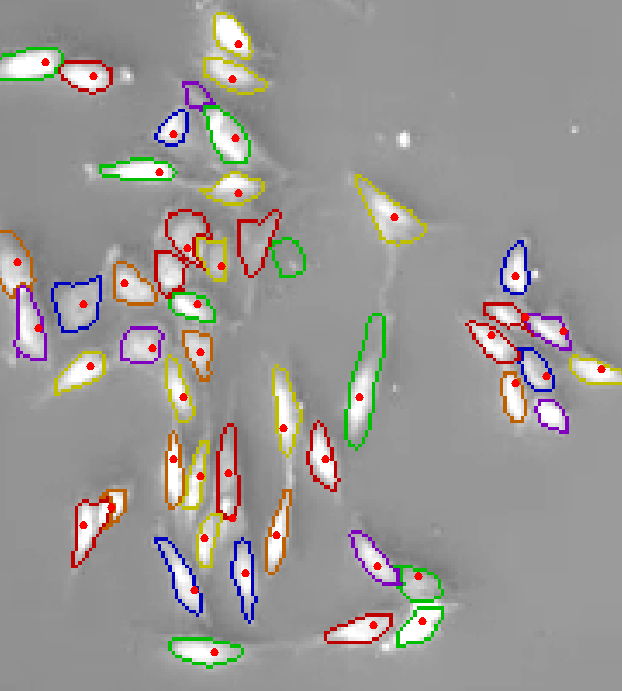}
    } 
 \caption{Datasets: A small region (200x200) from each dataset is shown along with the ground truth cell markers ({\color{red}$\bullet$}) and the boundaries of tracked cells.}
\label{fig:datasets}
\end{figure*}

We evaluate our proposed method on two fluorescence microscopy datasets, \textit{Fluo-N2DH-HeLa} and \textit{Fluo-N2DH-GOWT1}, and two phase contrast microscopy datasets, \textit{PhC-C2DH-U373} and \textit{PhC-C2DL-PSC} from ISBI 2015 Cell Tracking Challenge \cite{Maska2014}.
Each dataset contains 2 sequences in the training set; we train our method using one sequence and test on the other; this is repeated for both sequences in the dataset.
All these datasets only contain ground truth (GT) annotations for cells within a field of interest, which excludes few pixels wide border on all sides of images.
Two types of GT annotations are available: cell masks are annotated in few frames from each sequence and cell markers are annotated in the whole sequence except for \textit{PhC-C2DL-PSC} dataset which has only 101 frames annotated from each sequence.
Sample regions from each dataset along with the results are shown in Fig.~\ref{fig:datasets}.

\textbf{Fluo-N2DH-HeLa} contains fluorescently labeled HeLa nuclei.
Challenges with this dataset include high cell density, low contrast and few irregularly shaped cells.

\textbf{Fluo-N2DH-GOWT1} contains GFP transfected GOWT1 mouse embryonic stem cells.
The major challenge with this dataset is low contrast of some cells and few cells entering and exiting the imaged region from the axial direction.

\textbf{PhC-C2DH-U373} contains glioblastoma-astrocytoma U373 cells.
This dataset is challenging due to cells having highly deformable shapes and parts of cell bodies having similar appearance as the background.

\textbf{PhC-C2DL-PSC} contains pancreatic stem cells.
The challenges for this dataset include high cell density, elongated and deformable cell shapes, low inter-cellular shape and appearance variation and few very fast moving cells.

\subsection{Dataset Specific Adjustments}
We use the same network architecture, tracking graph model and training procedure for all four datasets except the dataset specific adjustments mentioned in this section.
The mean size of cell bounding box in the datasets varies considerably from 18 to 78.
Instead of modifying the network, we resize the images from datasets containing large cells, so that the network has enough spatial context to predict accurate cell bounding boxes and masks.
When proposing cell candidates, we rescale the \textit{PhC-C2DH-U373} and \textit{Fluo-N2DH-GOWT1} datasets so that the network's receptive field includes sufficient context to propose accurate bounding boxes and scores for most cells.
Both these datasets contain sparsely distributed cells so having little spatial context in case of larger than average cells is not a significant issue as the network is capable of proposing accurate boxes for objects which are larger than the receptive field of the network \cite{Rena}.

Median filtering is used to pre-process \textit{Fluo-N2DH-HeLa} and \textit{Fluo-N2DH-GOWT1}.
\textit{PhC-C2DH-U373} and \textit{PhC-C2DL-PSC} contain some GT markers very close to the cell boundary and small segmentation errors can result in the cells being counted as false negative.
We dilate the masks of tracked cells to mitigate miss-matches due to minor segmentation errors when evaluating tracking results.

For \textit{Fluo-N2DH-HeLa}, the limited training data (segmentation masks) available does not cover cell appearance and shape variation sufficiently for the network to learn the desired invariances especially when trained on less dense sequence and tested on dense sequence.
However, there are an order of magnitude more GT cell markers compared to GT masks.
We use these GT cell markers and marker controlled watershed to obtain cell segmentation and use this augmented training data to train our bounding box network.
This augmentation was necessary to achieve better performance than the \textit{BLOB} proposals \cite{Akram2016a}.

There are very few mitosis events in \textit{Fluo-N2DH-GOWT1} and \textit{PhC-C2DH-U373} datasets, so we disable mitosis edges for these datasets.
For \textit{PhC-C2DL-PSC} dataset, even though there are some mitosis events in the training data, we do not use these edges because the current mitosis detection module does not work very well.
This is partly due to these cells being very small, which results in failure to match proposals with some GT parent and daughters, which leads to some mitotic edges going to negative class and resulting in poor mitosis classifier performance.

\subsection{Baseline}
We compare our method against the best performing available methods for each dataset.
For fluorescent datasets, we use one tracking by detection (\textit{KTH} \cite{Magnusson2012a}) and three joint cell detection and tracking (\textit{EPFL} \cite{Engin2015}, \textit{HEID} \cite{Schiegg2014} and \textit{BLOB} \cite{Akram2016a}) methods as the baselines.
For \textit{PhC-C2DH-U373} dataset, we use the best performing \textit{U-Net} \cite{Ronneberger2015} and a graph cuts and model evolution based tracking method (\textit{GC-ME}) \cite{Bensch2015b} as the baselines.
For \textit{PhC-C2DL-PSC} dataset, there are no publicly available methods to our knowledge, so we only report our results.

\begin{table*}[!tbh]
\centering
\caption{Evaluation of proposals based tracking graphs ($G$) for all datasets.
Number of ground truth (GT) cell, and move and mitosis edges is listed along with their Recall (R).
\textit{R-NS} is the cell recall when cells matching under-segmented proposals are considered true positives (TP).}
\scalebox{1.2}{\begin{tabular}{@{\extracolsep{\fill}} l | l | c | c | c | c | c | c | c}
\hline\noalign{\smallskip}
\multicolumn{2}{c|}{} & \multicolumn{3}{|c|}{Cells} & \multicolumn{2}{|c}{Move Edges} & \multicolumn{2}{|c}{Mitosis Edges} \\
\cline{3-9}\noalign{\smallskip}
\multicolumn{2}{c|}{}	& 	GT & 	R		&	R-NS 	& 	GT &	R	&	GT & R	\\ 
\noalign{\smallskip}
\hline
\noalign{\smallskip}
\hline
\multirow{2}{*}{Fluo-N2DL-HeLa-01}
 & CPN &  \multirow{2}{*}{  8639}  & 0.99 & 1.00 &  \multirow{2}{*}{  8371}  & 0.99 &  \multirow{2}{*}{   188}  & 0.96\\ 
 & BLOB &  & 0.98 & 1.00 &  & 0.98 &  & 0.95\\ 
\hline
\multirow{2}{*}{Fluo-N2DL-HeLa-02}
 & CPN &  \multirow{2}{*}{ 25420}  & 0.98 & 1.00 &  \multirow{2}{*}{ 24730}  & 0.97 &  \multirow{2}{*}{   422}  & 0.92\\ 
 & BLOB &  & 0.98 & 1.00 &  & 0.97 &  & 0.93\\ 
\hline
\multirow{2}{*}{Fluo-N2DH-GOWT1-01}
 & CPN &  \multirow{2}{*}{  2052}  & 0.99 & 1.00 &  \multirow{2}{*}{  2021}  & 1.00 &  \multirow{2}{*}{     4}  & 0.00\\ 
 & BLOB &  & 0.98 & 0.99 &  & 0.97 &  & 0.00\\ 
\hline
\multirow{2}{*}{Fluo-N2DH-GOWT1-02}
 & CPN &  \multirow{2}{*}{  2321}  & 1.00 & 1.00 &  \multirow{2}{*}{  2270}  & 1.00 &  \multirow{2}{*}{     2}  & 0.00\\ 
 & BLOB &  & 1.00 & 1.00 &  & 1.00 &  & 0.00\\ 
\hline
\multirow{1}{*}{PhC-C2DH-U373-01}
 & CPN &  \multirow{1}{*}{   765}  & 1.00 & 1.00 &  \multirow{1}{*}{   757}  & 0.99 &  \multirow{1}{*}{     1}  & 0.00\\ 
\hline
\multirow{1}{*}{PhC-C2DH-U373-02}
 & CPN &  \multirow{1}{*}{   694}  & 0.84 & 0.93 &  \multirow{1}{*}{   679}  & 0.82 &  \multirow{1}{*}{     8}  & 0.00\\ 
\hline
\multirow{1}{*}{PhC-C2DL-PSC-01}
 & CPN &  \multirow{1}{*}{ 28687}  & 0.91 & 0.98 &  \multirow{1}{*}{ 28063}  & 0.89 &  \multirow{1}{*}{   426}  & 0.00\\ 
\hline
\multirow{1}{*}{PhC-C2DL-PSC-02}
 & CPN &  \multirow{1}{*}{ 23074}  & 0.89 & 0.99 &  \multirow{1}{*}{ 22542}  & 0.86 &  \multirow{1}{*}{   336}  & 0.00\\ 
\hline

\end{tabular}
}
\label{tab:graph_props}
\end{table*}
\newcommand{\iciplw}{.33\linewidth}
\begin{figure*}[!tbh]
  \centering
	\subfloat[Precision vs Recall (Markers)]{\label{fig:props_markers}\includegraphics[width=\iciplw]{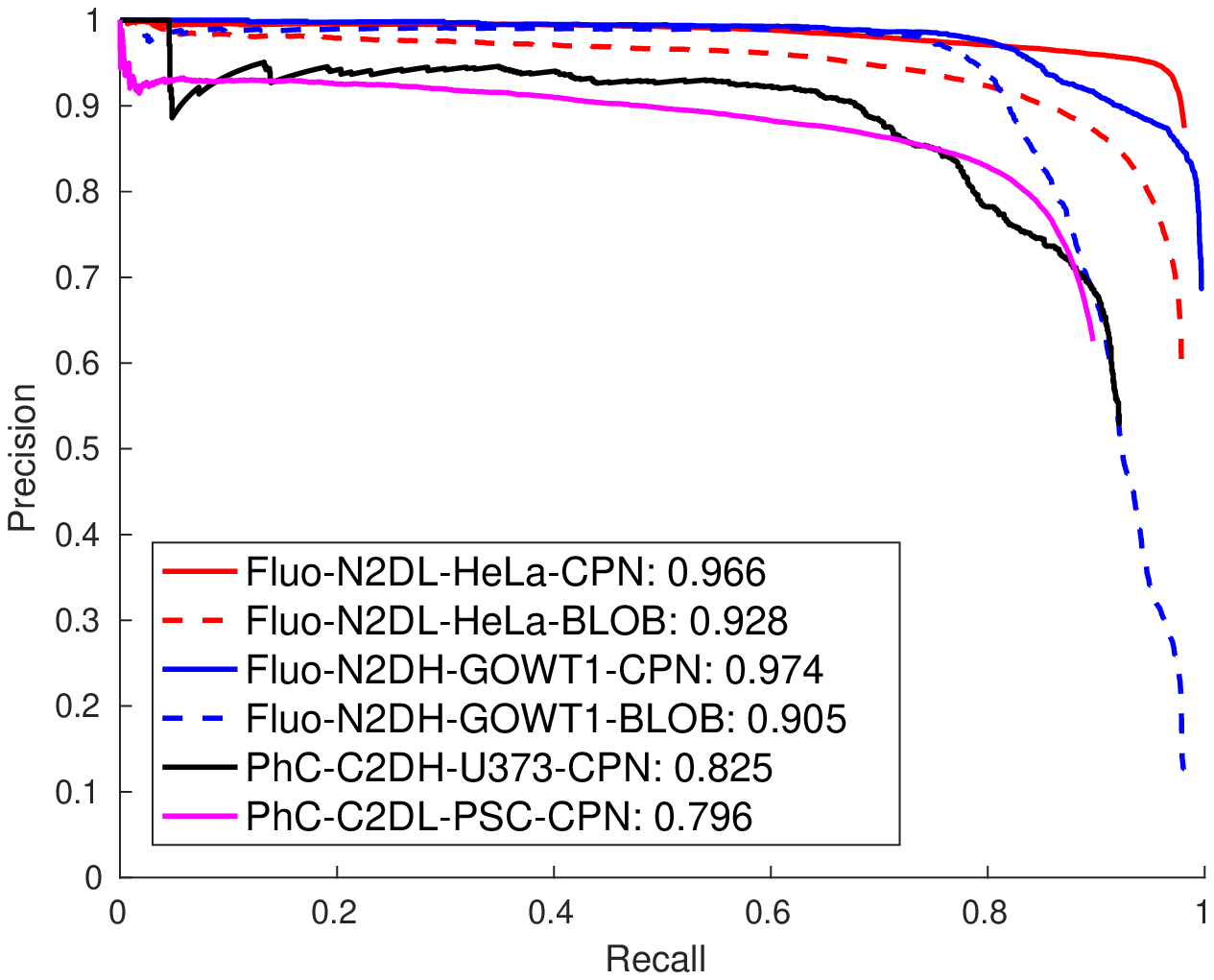}}
	\hfill
	\subfloat[Precision vs Recall (IoU)]{\label{fig:props_iou}\includegraphics[width=\iciplw]{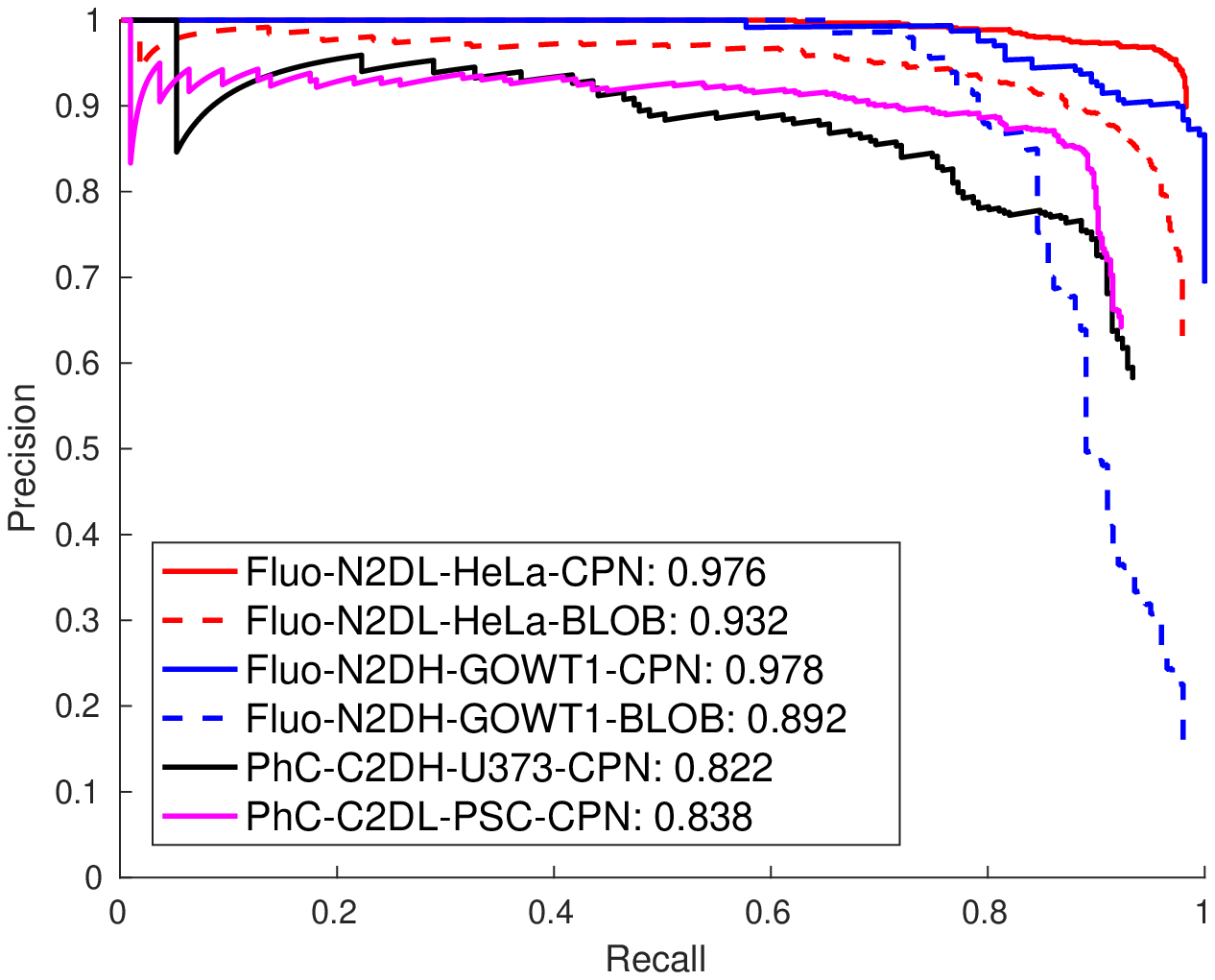}}
	\hfill
	\subfloat[Recall vs IoU]{\label{fig:props_ar}\includegraphics[width=\iciplw]{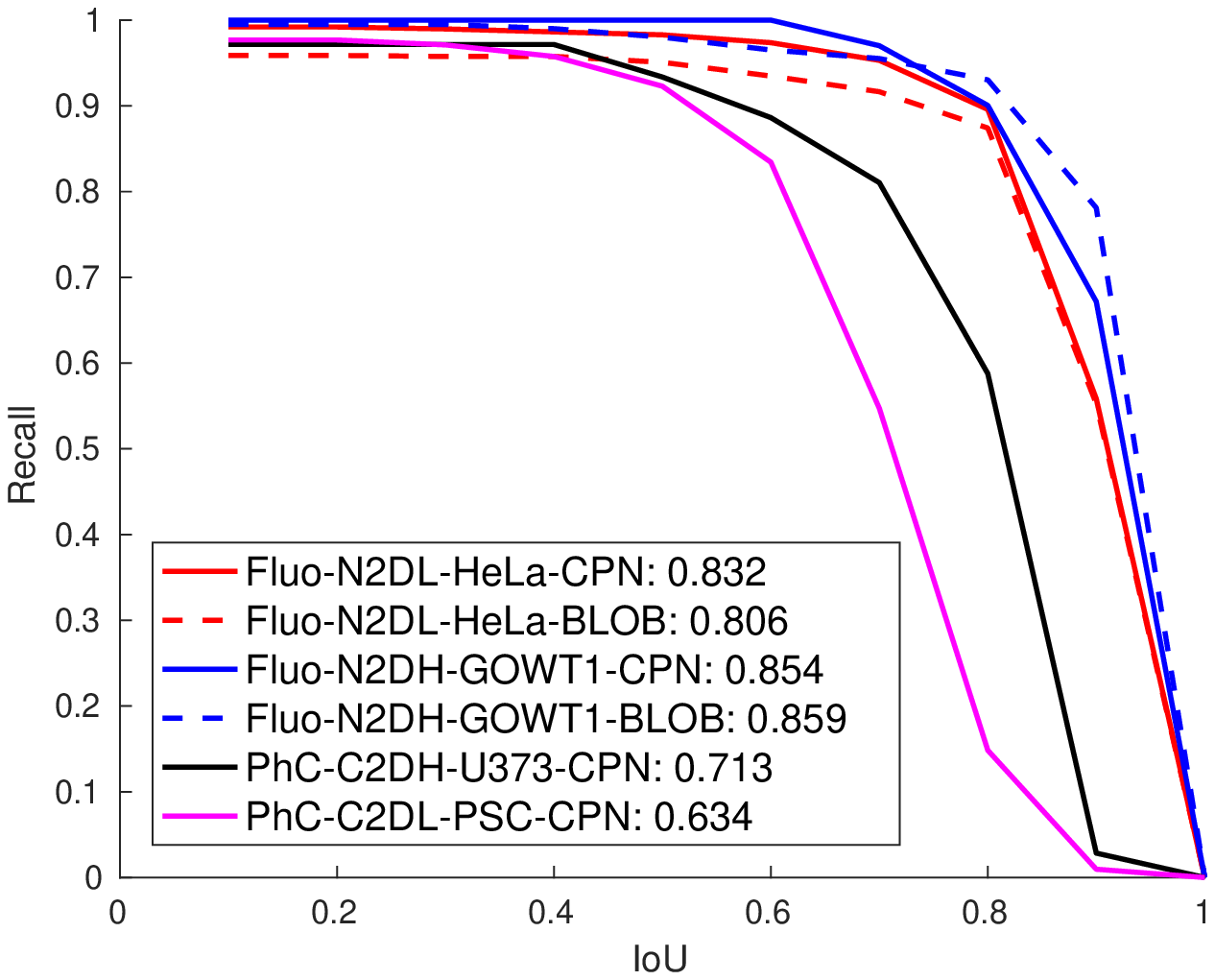}}
\caption{Comparison of BLOB and CPN proposals: Results in a) are computed using ground truth (GT) markers, while results in b) and c) are computed using GT masks.
Average precision (AP) values are in legend of a) and b), and average recall (AR) values are in legend of c).}
\label{fig:res_props}
\end{figure*}

\textbf{KTH} method, which won the ISBI cell tracking challenge \cite{Maska2014}, uses a band pass filter followed by thresholding to segment cells.
It uses watershed segmentation to split cell clusters and creates a tracking graph by connecting cell segmentations in adjacent frames.
It then finds cell tracks by iteratively finding the lowest cost path in this graph using Viterbi algorithm.

\textbf{EPFL} generates cell proposals by fitting ellipses to binary segmented regions and links proposals in adjacent frames in a tracking graph, which is solved using Integer linear programming (ILP).

\textbf{HEID} generates cell proposals by merging super-pixels, which are obtained using watershed.
Then a graphical model is used to represent cellular events and ILP is used to find the globally optimal tracking solution.

\textbf{BLOB} generates cell proposals by using multiple elliptical filter banks, and performs cell tracking by iteratively finding the shortest path in the tracking graph.

\textbf{U-Net} uses a U-shaped fully convolutional neural network to segment cells and grows cell tracks by linking segmentations in next frames with active tracks. 

\textbf{GC-ME} uses graph cuts with asymmetric boundary costs to segment cells and uses the segmentation from previous frame to segment new frame and grow cell tracks.

For EPFL and HEID, we report the results provided by \cite{Engin2015} and for KTH, we report the results of the provided tracking software\footnote{\url{http://www.codesolorzano.com/Challenges/CTC/KTH-SE_2013.html}}.
For U-Net\footnote{\url{http://lmb.informatik.uni-freiburg.de/people/ronneber/u-net/}} and GC-ME\footnote{\url{http://lmb.informatik.uni-freiburg.de/Publications/2015/BR15/}}, we report results of the trained models provided along with their codes.

\subsection{Results}

\subsubsection{Proposal Graph}
Table~\ref{tab:graph_props} lists the number of ground truth (GT) cells, and move and mitosis edges in each sequence along with their recall (R) in the proposal tracking graph.
Recall of individual cells (R) considers a GT cell found if there is a proposal which only matches this cell.
\textit{R-NS} is the cell recall, which does not penalize the under-segmented proposals, i.e. if a proposal matches multiple GT cells, these GT cells are considered found.
If a proposal is found for the GT cells on both sides of a GT edge and those proposals are linked with the edge of same class (mitosis, move) then the edge is considered found.

\textit{CPN} graphs have slightly higher recall for cells, move and mitosis edges compared to \textit{BLOB} graphs.
Cell recall for \textit{U373-02} sequence is low mostly due to some GT cells having their markers very close to cell boundary, and slight errors in proposal masks result in miss-matches between the proposals and the GT markers.
For \textit{PhC-C2DL-PSC} sequences, cell recall is also quite low but the \textit{R-NS} is nevertheless quite high; this is because \textit{PSC} cells are small, thin and elongated, and cell density in these sequences is high.
As a result minor segmentation errors can lead to a proposal either missing a GT marker or matching multiple GT markers.
Bounding boxes are also not a very good representation for these cells.
Two or more cell proposal bounding boxes can have very large overlap with each other and during non-maxima suppression, some good proposal may be deleted.

Almost all of the false negative (FN) move edges are due to FN cells.
\textit{PSC} sequences contain some cells which move very rapidly and these cells contribute some FN move edges due to the gating used when associating proposals between adjacent frames.

\subsubsection{Cell Proposals}
We use average precision (AP) - area under precision-recall curve - and average recall (AR) - area under recall-IoU curve to evaluate cell proposals.
We match cell proposals with GT cells using two different criteria.
First criteria utilizes GT segmentation masks; a proposal is true positive (TP) if it has an IoU $>$ 0.5 with any unmatched GT cell, otherwise it is false positive (FP).
Second criteria utilizes GT markers; a proposal is TP if it has only one marker, which is unmatched, inside its mask, otherwise it is FP.
GT markers/masks which remain unmatched are counted as false negatives (FN).
Cell proposals from all frames from both sequences in a dataset are combined and ranked by their score and then evaluated as TP or FP.
This provides a pair of recall (R = TP/(TP+FN)) and precision (P = TP/(TP+FP)) values after evaluating each proposals, which are reported in the precision-recall curves of Fig.~\ref{fig:res_props}.

\textit{CPN} proposals consistently have better precision and higher average precision values for both datasets, indicating that their scoring is much better than \textit{BLOB} proposals which rely on hand crafted features.
\textit{CPN} proposals also have slightly higher final recall than \textit{BLOB} proposals.
\textit{CPN} proposals have higher recall for low intersection over union overlap (IoU) values but the difference between \textit{BLOB} and \textit{CPN} proposals decreases as IoU threshold increases, with \textit{BLOB} achieving slightly higher recall for IoU thresholds above 0.8 on \textit{GOWT1} dataset.
There are three factors for this lower recall at high IoU values; firstly, some bounding boxes proposed by \textit{CPN} are not very well localized and parts of cells outside these boxes can not be in the segmented masks; secondly, the cell proposal network uses lower resolution images; thirdly, the output masks from the network have fixed size which is much smaller than the mean cell size.


\subsubsection{Edge Proposals}
Fig.~\ref{fig:res_edge_props} shows the precision-recall curve for move and mitosis edges.
Even though same features and classifiers are used for scoring move and mitosis edges in the \textit{BLOB} and \textit{CPN} graphs, there is considerable difference between their performance due to the quality of cell proposals and it highlights the importance of good cell proposals even in the scoring of edges in the tracking graph.
A bad proposal results in additional confounding edges and increases the chances of selection of a bad move/mitosis edge.
Mitosis events are quite rare and there are very limited mitosis samples in the sequences, which is one of the factor for relatively poor mitosis classifier performance.
Nevertheless, it is good enough for detecting most mitosis events during tracking when temporal information is also utilized.

\subsubsection{Mitosis}
We use F1-Score ($F1 = 2 \cdot \frac{P \cdot R}{P+R}$), harmonic mean of precision (P) and recall (R), to compare mitosis detection performance.
Table~\ref{tab:mitosis} lists the F1-Score, recall and precision of mitosis detection in the tracking results for all methods on \textit{Fluo-N2DL-HeLa} dataset.
Our method has the second highest recall and F1-score behind \textit{EPFL} method.
Daughter cells can be very small and when they are very close to much bigger cells, our network sometimes fails to propose candidates for these daughter cells and misses the mitosis event.


\begin{figure}[tbh]
  \centering
	\subfloat[Move edges]{\label{fig:props_move}\includegraphics[width=0.5\linewidth]{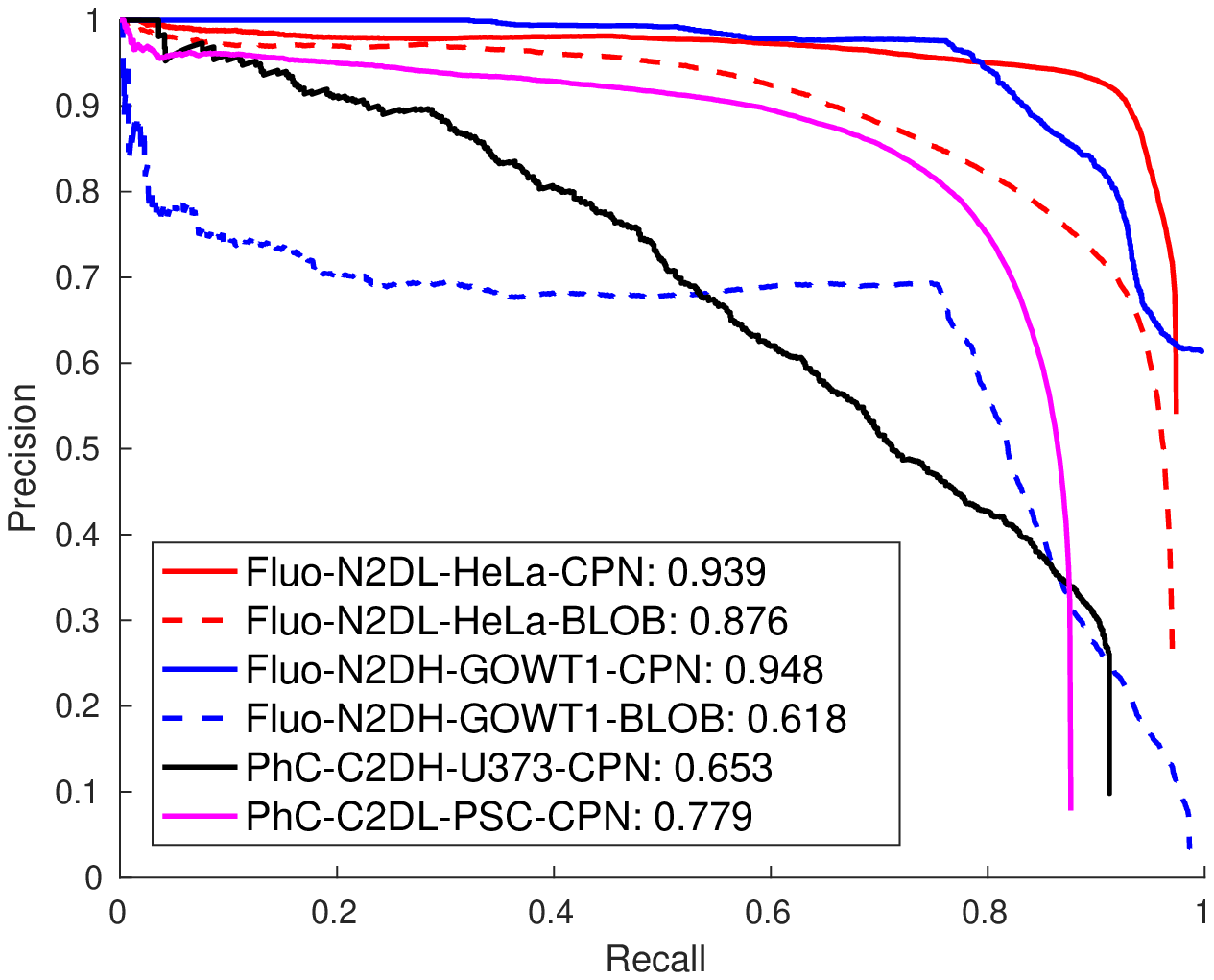}}
	\subfloat[Mitosis edges]{\label{fig:props_mitosis}\includegraphics[width=0.5\linewidth]{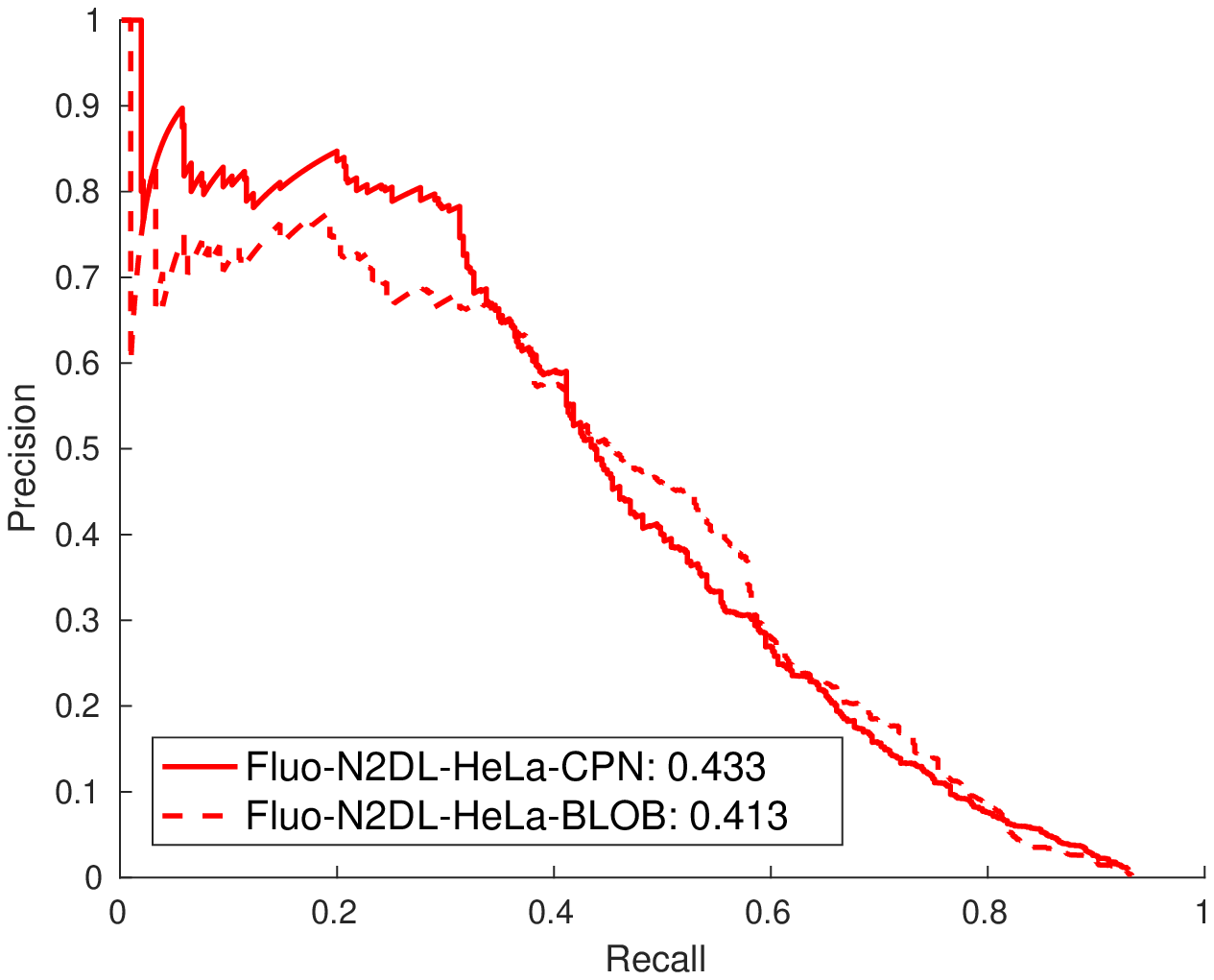}}
\caption{Evaluation of move and mitosis edge proposals. Average precision (AP) is shown in the legend.}
\label{fig:res_edge_props}
\end{figure}

\begin{table}[tbh]
\centering
\caption{Mitosis detection results for our method (CPN), BLOB\cite{Akram2016a}, KTH \cite{Magnusson2012a}, EPFL \cite{Engin2015}, HEID \cite{Schiegg2014}.
\textsuperscript{*}: Results are from \cite{Engin2015}.}
\scalebox{1.2}{\begin{tabular}{@{\extracolsep{\fill}} l | l | c | c | c }
\hline\noalign{\smallskip}
    						&				& 	F1		&	Recall 	& 		Precision	\\ 
\noalign{\smallskip}
\hline
\noalign{\smallskip}
\hline
\multirow{5}{*}{Fluo-N2DL-HeLa-01}
 &        CPN & \textbf{0.86} & 0.82 & 0.91\\ 
 &       BLOB & 0.61 & 0.61 & 0.61\\ 
 &        KTH & 0.66 & 0.74 & 0.59\\ 
 &       EPFL\textsuperscript{*} & 0.85 & 0.92 & 0.79\\ 
 &       HEID\textsuperscript{*} & 0.65 & 0.79 & 0.55\\ 
\hline
\multirow{5}{*}{Fluo-N2DL-HeLa-02}
 &        CPN & 0.76 & 0.71 & 0.81\\ 
 &       BLOB & 0.54 & 0.70 & 0.44\\ 
 &        KTH & 0.56 & 0.66 & 0.49\\ 
 &       EPFL\textsuperscript{*} & \textbf{0.84} & 0.86 & 0.83\\ 
 &       HEID\textsuperscript{*} & 0.54 & 0.69 & 0.44\\ 
\hline

\end{tabular}
}
\label{tab:mitosis}
\end{table}

\begin{table*}[!tbh]
\centering
\caption{Tracking (TRA) and segmentation (SEG) performance for our method (CPN), KTH \cite{Magnusson2012a}, EPFL \cite{Engin2015}, HEID \cite{Schiegg2014}, BLOB \cite{Akram2016a}, U-Net \cite{Ronneberger2015} and GC-ME \cite{Bensch2015b}.
Number of different types of errors (FN: False negative cells, NS: Under-segmented regions, EA: False negative edge, EC: Wrongly labeled edges and ED2: False positive edges) in the tracking results are listed as well.
Best TRA and SEG values for each sequence are highlighted.
\textsuperscript{*} denotes results from \cite{Engin2015} and \textsuperscript{$\dagger$} denotes results for models trained using data from both sequences.}
\scalebox{1.2}{\begin{tabular}{@{\extracolsep{\fill}} l | l | c | c | c | c | c | c | c | c }
\hline\noalign{\smallskip}
    						&				& 	TRA		&	SEG 	& 		FN	&	FP	&	NS	&	EA 	&	EC	&	ED2 	\\ 
\noalign{\smallskip}
\hline
\noalign{\smallskip}
\hline
\multirow{5}{*}{Fluo-N2DL-HeLa-01}
 &        CPN & \textbf{0.9869} & \textbf{0.8313} &    48 &   247 &    65 &   139 &    23 &    21 \\ 
 &       BLOB & 0.9803 & 0.7951 &    82 &   359 &    81 &   175 &    74 &    39 \\ 
 &        KTH & 0.9775 & 0.8018 &   143 &   318 &    17 &   222 &    38 &    27 \\ 
 &       EPFL\textsuperscript{*} & 0.98 &       &       &       &       &       &       &       \\ 
 &       HEID\textsuperscript{*} & 0.80 &       &       &       &       &       &       &       \\ 
\hline
\multirow{5}{*}{Fluo-N2DL-HeLa-02}
 &        CPN & \textbf{0.9826} & \textbf{0.8445} &   155 &  1288 &   281 &   470 &    79 &    54 \\ 
 &       BLOB & 0.9771 & 0.8390 &   198 &  1423 &   363 &   655 &   271 &   209 \\ 
 &        KTH & 0.9747 & 0.8366 &   373 &  1146 &   269 &   633 &   141 &    80 \\ 
 &       EPFL\textsuperscript{*} & 0.97 &       &       &       &       &       &       &       \\ 
 &       HEID\textsuperscript{*} & 0.85 &       &       &       &       &       &       &       \\ 
\hline
\multirow{5}{*}{Fluo-N2DH-GOWT1-01}
 &        CPN & \textbf{0.9864} & \textbf{0.8506} &    15 &    84 &     0 &    57 &     1 &     1 \\ 
 &       BLOB & 0.9733 & 0.7415 &    41 &   121 &     3 &    56 &     0 &     0 \\ 
 &        KTH & 0.9462 & 0.6849 &    97 &   147 &     0 &   100 &     0 &     0 \\ 
\hline
\multirow{5}{*}{Fluo-N2DH-GOWT1-02}
 &        CPN & \textbf{0.9719} & 0.8725 &    12 &   606 &     0 &    13 &     0 &     3 \\ 
 &       BLOB & 0.9628 & \textbf{0.9046} &    46 &   477 &     0 &    35 &     1 &     0 \\ 
 &        KTH & 0.9452 & 0.8942 &   121 &    90 &     0 &   106 &     1 &     0 \\ 
 &       EPFL\textsuperscript{*} & 0.95 &       &       &       &       &       &       &       \\ 
 &       HEID\textsuperscript{*} & 0.95 &       &       &       &       &       &       &       \\ 
\hline
\multirow{3}{*}{PhC-C2DH-U373-01}
 &        CPN & 0.9594 & 0.7336 &     1 &   288 &     3 &    28 &     0 &     2 \\ 
 &      U-Net\textsuperscript{$\dagger$} & \textbf{0.9869} & \textbf{0.9375} &     0 &    12 &    17 &    12 &     0 &     0 \\ 
 &      GC-ME\textsuperscript{$\dagger$} & 0.9779 & 0.8748 &     2 &   167 &     0 &     5 &     0 &     0 \\ 
\hline
\multirow{3}{*}{PhC-C2DH-U373-02}
 &        CPN & 0.9346 & 0.7376 &    26 &    25 &    37 &    33 &     1 &     1 \\ 
 &      U-Net\textsuperscript{$\dagger$} & \textbf{0.9547} & \textbf{0.8303} &     3 &    26 &    57 &    12 &     1 &     1 \\ 
 &      GC-ME\textsuperscript{$\dagger$} & 0.9040 & 0.7567 &    37 &   121 &    36 &    63 &     0 &     0 \\ 
\hline
\multirow{1}{*}{PhC-C2DL-PSC-01}
 &        CPN & \textbf{0.9429} & \textbf{0.6606} &   427 &  3273 &  1777 &  1459 &   183 &    19 \\ 
\hline
\multirow{1}{*}{PhC-C2DL-PSC-02}
 &        CPN & \textbf{0.9358} & \textbf{0.6476} &   420 &  1682 &  1761 &  1443 &   135 &    34 \\ 
\hline

\end{tabular}
}
\label{tab:results}
\end{table*}

\subsubsection{Tracking}
We use the tracking performance measure (TRA) \cite{Matula2015} and Jaccard similarity index (SEG) used in the ISBI 2015 Cell Tracking Challenge for evaluating the tracking results.
TRA is designed to mirror the manual effort required to correct the tracking errors generated by a tracking method and it penalizes the following errors: FN (False Negatives), FP (False Positives), NS (Under-segmentations), EA (Missing edges), EC (Miss-labeled edges) and ED2 (False positive edges).
Both TRA and SEG values range between 0 and 1 (perfect result).

Table~\ref{tab:results} lists the results for our method (\textit{CPN}) and compares them with all baseline methods.
Our method achieves better tracking score, higher TRA, on all four fluorescent sequences.
It consistently achieves significantly lower false negative (FN) and association errors (EA, EC, ED2).
It has higher false positive (FP) errors; this is partly due to the presence of some very dark and low contrast cell like regions in fluorescent datasets, which resemble cells but are not annotated as cells and some over-segmentation errors.

For \textit{PhC-C2DH-U373} sequences our method has lower performance than \textit{U-Net}.
One reason for lower performance of our method is the fact that \textit{U-Net} model is trained using data from both sequences, while we use data from only one sequence to train our models and there are some cell shapes in both sequences which are not covered by the data in other sequence and deformation or other augmentations are not sufficient to learn these cell shapes/appearances.
It is also clear from Table~\ref{tab:results} that our method and \textit{U-Net} have different types of errors; our method has fewer under-segmentation (NS) errors but many more FP errors.
Most of these FPs are due to over-segmentation of cells which are much bigger than the average cell and the network neither has enough spatial context nor has it seen enough training samples to propose accurate bounding boxes.
\textit{U-Net} produces very accurate cell masks but when cells come in contact with each other, it often fails to separate them.
Our method is able to propose good candidates for most individual cells but its segmentation masks are very coarse.

\textit{PhC-C2DL-PSC} sequences are the most challenging due to cells being small and elongated, with very similar appearances.
Many of the errors (FN/FP/NS) in these sequences are due to minor segmentation errors, which result in failure to match the detected cells with the GT cells.

\fi

\section{Discussion}
\label{sec:disc}
The recent developments in microscopic imaging have opened up possibilities for quantitative analyses of image data, including analysis of large number of single cells after successful cell segmentation and tracking in time-lapse images.
One of the most important challenge when developing a general cell tracking method is the huge variability in the cell shapes and appearances.
Cell tracking literature is filled with methods \cite{Khan2014,Stegmaier2016}, which are designed for a narrow set of sequences by fine-tuning a chain of simple image processing operations to answer the biologically relevant question.
These methods can achieve very good performance on the specific sequences they are designed for but it can require substantial effort to tune their parameters when analyzing other sequences.
To make significant progress in cell tracking field, it is crucial that the cell tracking methods should be able to cope with a wider range of sequences.

In this paper, we have presented a simple and completely automated cell tracking method, which represents multiple hypothesis for ambiguous regions using cell proposals; links cell proposals in adjacent frames using edges representing cellular events; and uses Integer Linear Programming to select the optimal subset of cell and edge proposals to obtain accurate and robust cell tracks.
Our method has excellent overall tracking performance as indicated in Table~\ref{tab:results} and it is the only currently available method that is directly applicable to all of our challenging image sequences and which achieves good performance without laborious manual hand-tuning.


There are some components of our tracking method which have room for improvement.
Firstly, our mitosis detector is not very general and may not work well on new sequences which do not satisfy the underlying assumptions.
This task is often further complicated by the limited number of mitosis events in the training data.
Secondly, our tracking graph (as is common for most cell tracking methods \cite{Engin2015,Magnusson2015}) does not use any motion model.
It is not easy to incorporate cell motion model directly in the current formulation but cell speed cues can however be incorporated using optical flow \cite{Amat2013a} or correlation matching \cite{Schiegg2013} and for some cell tracking applications, e.g. scratch assay for cell migration \cite{Kaakinen2014}, they may improve performance.

The masks predicted by our network are coarse for large cells and there is considerable room for improvement.
One reason for the coarse masks is the use of fully connected layers which limits the resolution of predicted masks.
Combining the output of this fully connected layer with the output of a parallel branch containing convolutional layer which predicts a mask of same size as the candidate bounding box can potentially result in some improvement.

The recall of our proposal network is slightly lower for small and elongated cells compared to compact average sized cells.
Use of feature pyramids \cite{Lin2016} and hard mining \cite{Shrivastava2016} are some of the extensions which may improve the recall for these difficult cells.


Analysis of errors in the proposal graph indicates that the under-segmentation errors are one major bottleneck in the performance of our method.
In many of these regions, information in a single image can be misleading and it can be very challenging to include correct cell candidates without growing the number of total proposals considerably.
One potential solution that can mitigate this issue is to allow multiple cells in each proposal, which may improve final tracking performance.

Our cell proposal generation network produces very accurate cell scores and can achieve very high recall with a small set of cell candidates.
This drastically reduces the number of move and mitosis edges in the tracking graph \cite{Akram2016a,Engin2015,Schiegg2014} and leads to much faster tracking with optimization taking few minutes for \textit{PhC-C2DL-PSC-01} sequence containing $\approx$ 175,000 cells.

\section{Conclusion}
\label{sec:conc}
In this paper, we have presented a cell proposal based joint cell detection and tracking method, which performs cell tracking by selecting an optimal subset of cell and edge proposals.
Our method is completely automated and given sufficient training data, it can be applied to sequences of other cell types and microscopy modalities.
We have compared our method against state of the art cell tracking methods and shown that it outperforms them on multiple fluorescence and phase contrast microscopy sequences.
Code is available at: \href{https://github.com/SaadUllahAkram/CellTracker}{https://github.com/SaadUllahAkram/CellTracker}.
\FloatBarrier
\printbibliography

\end{document}